\documentclass[10pt,twocolumn,letterpaper]{article}

\DeclareMathAlphabet\mathbfcal{OMS}{cmsy}{b}{n}

\usepackage{cvpr}
\usepackage{times}
\usepackage{epsfig}
\usepackage{graphicx}
\usepackage{amsmath}
\usepackage{amssymb}

\usepackage{array}
\usepackage{multirow}
\usepackage{diagbox}
\usepackage{tabularx}
\usepackage{subfig}

\usepackage[pagebackref=true,breaklinks=true,letterpaper=true,colorlinks,bookmarks=false]{hyperref}

\cvprfinalcopy 

\def\cvprPaperID{5234} 

\ifcvprfinal\fi
\begin{document}

\title{Semi-Supervised Semantic Image Segmentation with Self-correcting Networks}

\author{Mostafa S. Ibrahim\thanks{Work done while interning at D-Wave Systems} \\
Simon Fraser University \\ 
{\tt\footnotesize msibrahi@sfu.ca}
\and
Arash Vahdat \\
NVIDIA \\ 
{\tt\footnotesize avahdat@nvidia.com}
\and
Mani Ranjbar \\
Sportlogiq\\ 
{\tt\footnotesize mani@sportlogiq.com}
\and
William G. Macready \\
Sanctuary AI \\ 
{\tt\footnotesize wgm@sanctuary.ai}
}

\maketitle

\renewcommand{\ll}         {\pmb l}
\newcommand{\x}         {\pmb x}
\newcommand{\xf}         {\x^{(f)}}
\newcommand{\xw}         {\x^{(w)}}
\newcommand{\y}         {\pmb y}
\newcommand{\yf}         {\y^{(f)}}
\newcommand{\yw}         {\y^{(w)}}
\newcommand{\bb}        {\pmb b}
\newcommand{\bbf}       {\bb^{(f)}}
\newcommand{\bbw}       {\bb^{(w)}}
\newcommand{\F}         {\mathbfcal{F}}
\newcommand{\W}         {\mathbfcal{W}}
\newcommand{\pest}      {$p_{anc}   (\pmb y|\pmb x, \pmb b)$ \hspace{0.5mm}}
\newcommand{\psegnet}   {$p  (\pmb y|\pmb x)$ \hspace{0.5mm}}  

\newcommand{\bphi}      {\pmb \phi}
\newcommand{\btheta}    {\pmb \theta}
\newcommand{\blambda}   {\pmb \lambda}

\newcommand{\palphacomp}{$p_{\alpha}(\pmb y|\pmb x)$ \hspace{1mm}}
\newcommand{\pnetcomp}  {$p_{net}  (\pmb y|\pmb x)$ \hspace{1mm}}
\newcommand{\KL}        {\text{KL}}
\newcommand{\arash}[1]  {{\color{red} #1}}

\newcommand{\photo}[1]{%
    \includegraphics[width=2.75cm, height=1.75cm]{#1}
}

\captionsetup{skip=6.5pt}

\begin{abstract}
Building a large image dataset with high-quality object masks for semantic segmentation is costly and time consuming. In this paper, we introduce a principled semi-supervised framework that only uses a small set of fully supervised images (having semantic segmentation labels and box labels) and a set of images with only object bounding box labels (we call it the weak set). Our framework trains the primary segmentation model with the aid of an ancillary model that generates initial segmentation labels for the weak set and a self-correction module that improves the generated labels during training using the increasingly accurate primary model. We introduce two variants of the self-correction module using either linear or convolutional functions. Experiments on the PASCAL VOC 2012 and Cityscape datasets show that our models trained with a small fully supervised set perform similar to, or better than, models trained with a large fully supervised set while requiring $\sim$7x less annotation effort.
\end{abstract}

\vspace{-0.5cm}
\section{Introduction}
Deep convolutional neural networks (CNNs) have been 
successful in many computer vision tasks including image
classification~\cite{KrizhevskyNIPS12AlexNet, HeCVPR16ResNet, ZophCVPR18NASNet}, object detection~\cite{RenNIPS15FasterRCNN, LiuECCV16SSD, RedmonCVPR16YOLO}, semantic segmentation~\cite{BadrinarayananArXiv15SegNet, ZhaoCVPR17PyramidParsing, ChenECCV18Deeplabv3plus}, 
action recognition~\cite{DonahueCVPR15LSTM, KarpathyCVPR14CNNVideo, SimonyanNIPS14Action, TranCVPR15conv3D}, and facial landmark localization~\cite{SunCVPR13Face, ZhangECCV14facial, ZhuCVPR16face}. However, the
common prerequisite for all these successes is the 
availability of large training corpora of labeled images. 
Of these tasks, semantic image segmentation
is one of the most costly tasks in terms of data annotation.
For example,  drawing 
a segmentation annotation on an object is on average $\sim$8x slower 
than drawing a bounding box and $\sim$78x slower than
labeling the presence of objects in images~\cite{BearmanECCV16WeakPoint}. 
As a result, most image segmentation
datasets are orders of magnitude smaller than image-classification datasets.

In this paper, we mitigate the data demands of semantic segmentation with a semi-supervised method that leverages cheap object bounding box labels in training. This approach reduces the data annotation requirements at the cost of requiring inference of the mask label for an object within a bounding box.

Current state-of-the-art semi-supervised methods 
typically rely on hand-crafted heuristics to infer an object mask inside 
a bounding box~\cite{PapandreouICCV15ExpMax, DaiICCV15BoxSup, KhorevaCVPR17SimpleDoes}. 
In contrast, we propose a principled framework that trains semantic segmentation models in a semi-supervised 
setting using a small set of fully supervised images (with semantic object masks and bounding boxes) and a weak set of images (with only bounding box annotations). The fully supervised set is 
first used to train an ancillary segmentation model that predicts object masks on the weak set.
Using this augmented data a primary segmentation model is trained. This primary segmentation model is probabilistic to accommodate the uncertainty of the mask labels generated by the ancillary model.
Training is formulated so that the labels supplied to the primary model are refined during training from the initial ancillary mask labels to more accurate labels obtained from the primary model itself as it improves.
Hence, we call our framework a self-correcting segmentation model as it improves the weakly supervised labels based on its current probabilistic model of object masks.

We propose two approaches to the self-correction mechanism.
Firstly, inspired by Vahdat~\cite{VahdatNIPS17Robust}, we use a function that linearly combines
the ancillary and model predictions.
We show that this simple and effective approach
is the natural result of minimizing a weighted Kullback-Leibler (KL) divergence from
a distribution over segmentation labels to both the ancillary and primary models.
However, this approach requires defining a weight whose optimal value should
change during training. With this motivation, we develop a second adaptive self-correction mechanism.
We use CNNs to learn how to combine the ancillary and primary models to predict a segmentation on a weak set of images. This approach eliminates the need for a weighting schedule.

Experiments on the PASCAL VOC and Cityscapes datasets show that 
our models trained with a small portion of fully supervised set achieve a
performance comparable to (and in some cases better than) the models 
trained with all the fully supervised images.


\section{Related Work}
\vspace{-0.1cm}
\paragraph{Semantic Segmentation:} Fully convolutional networks (FCNs)~\cite{LongCVPR15SegFully}
have become indispensable models for semantic image segmentation. Many successful
applications of FCNs rely on atrous convolutions~\cite{YuICLR15SegMultiScale} (to increase the receptive field of the network without down-scaling the image) and dense conditional random fields (CRFs)~\cite{KrahenbuhlNIPS11DenseCRF} (either as post-processing~\cite{ChenICLR15DeeplabV1} or as an integral part of the segmentation model~\cite{ZhengCVPR15CRFSegmentation, LinCVPR16piecewise, Schwing2015fully, LiuICCV15DeepParsing}).
Recent efforts have focused on encoder-decoder based models that extract long-range information
using encoder networks whose output is passed to decoder networks that generate 
a high-resolution segmentation prediction. SegNet~\cite{BadrinarayananArXiv15SegNet},
U-Net~\cite{RonnebergerMICCAI15UNet} and RefineNet~\cite{LinCVPR17RefineNet} 
are examples of such models that use different mechanisms for passing information from the encoder to the decoder.\footnote{SegNet~\cite{BadrinarayananArXiv15SegNet} transfers max-pooling indices from encoder to decoder, \mbox{U-Net}~\cite{RonnebergerMICCAI15UNet} introduces skip-connections between encoder-decoder networks and RefineNet~\cite{LinCVPR17RefineNet} proposes multipath refinement in the decoder through long-range residual blocks.}
Another approach for capturing long-range contextual information is spatial pyramid
pooling~\cite{LazebnikCVPR06SPM}. ParseNet~\cite{Liu15Parsenet} adds global context
features to the spatial features, DeepLabv2~\cite{ChentPAMI17DeeplabV2} uses
atrous spatial pyramid pooling (ASPP), and 
PSPNet~\cite{ZhaoCVPR17PyramidParsing} introduces
spatial pyramid pooling on several scales for the segmentation problem.

While other segmentation models may be used, we employ DeepLabv3+~\cite{ChenECCV18Deeplabv3plus} as our segmentation model because it outperforms previous CRF-based DeepLab models using simple factorial output. DeepLabv3+ replaces Deeplabv3's \cite{ChentArXiv17DeeplabV3} backbone with the Xception network~\cite{CholletCVPR17Xception} and stacks it with a simple two-level decoder that uses lower-resolution feature maps of the encoder. 

\vspace{-0.3cm}
\paragraph{Robust Training:} Training a segmentation model from bounding box information
can be formulated as a problem of robust learning from
noisy labeled instances. 
Previous work on robust learning has focused on classification problems with a small number of output variables. In this setting, a common simplifying assumption models the noise on output labels
as independent of the input~\cite{NatarajanNIPS13noisy, MnihICML12Arial, PatriniCVPR17, Sukhbaatar14Noisy, ZhangNIPS18generalized}.
However, recent work has lifted this constraint to model noise based on each instance's content (i.e., input-dependent noise). Xiao \etal~\cite{XiaoCVPR15} use a simple binary indicator function to represent whether each instance does or does not have a noisy label. Misra \etal~\cite{MisraCVPR16LabelingBias} represent label noise for each class independently.
Vahdat~\cite{VahdatNIPS17Robust} proposes CRFs to represent the joint distribution of noisy and clean labels extending structural models \cite{VahdatM13,VahdatZM14} to deep networks. Ren \etal~\cite{RenICML18Robust} gain robustness against noisy labels by reweighting each instance during training whereas Dehghani \etal~\cite{DehghaniICLR18fidelity} reweight gradients based on a confidence score on labels.
Among methods proposed for label correction, Veit \etal~\cite{VeitCVPR17Noisy} use a neural regression model to predict clean labels given noisy labels and image features, Jiang \etal~\cite{JiangICML18mentornet} learn curriculum, and Tanaka \etal~\cite{TanakaCVPR18Noisy} use the current model to predict labels. All these models have been restricted to image-classification problems and have not yet been applied to image segmentation.
\vspace{-0.3cm}
\paragraph{Semi-Supervised Semantic Segmentation:} The focus of this paper
is to train deep segmentation CNNs using bounding box annotations.
Papandreou \etal~\cite{PapandreouICCV15ExpMax} propose an Expectation-Maximization-based (EM) algorithm on top of DeepLabv1~\cite{ChenICLR15DeeplabV1} to estimate segmentation labels for the weak set of images (only with box information). In each training step, segmentation labels are estimated based on the network output in an EM fashion. Dai \etal~\cite{DaiICCV15BoxSup} propose an iterative training approach that alternates between generating region proposals (from a pool of fixed proposals) and fine-tuning the network. Similarly, Khoreva \etal~\cite{KhorevaCVPR17SimpleDoes} use an iterative
algorithm but rely on GrabCut~\cite{RotherACM04grabcut} and hand-crafted rules to extract the segmentation mask in each iteration. Our work differs from these previous methods in two significant aspects: i) We replace hand-crafted rules with an ancillary CNN for extracting 
probabilistic segmentation labels for an object within a box for the weak set. ii) We use a self-correcting model to correct for the mismatch
between the output of the ancillary CNN and the primary segmentation model during training.

In addition to box annotations, segmentation models may use other forms of weak annotations such as image pixel-level~\cite{WangCVPR18WeakImgLevelMining, WeiCVPR18WeakImgLevelDilated, HuangCVPR188WeakImgLevelSeeded, AhnCVPR18WeakImgLevelAffinity, GeCVPR18WeakImgLevelEvidence, WeiCVPR18WeakImgLevelAdversarialErasing, DurandCVPR18WeakImgLevelWILDCAT}, image label-level ~\cite{ZhangCVPR15SegWeakSocial}, scribbles~\cite{XuCVPR15Scribble, LinCVPR16ScribbleSup}, point annotation~\cite{BearmanECCV16WeakPoint}, or web videos~\cite{HongCVPR17SegWeakWeb}.
Recently, adversarial learning-based methods~\cite{HungBMVC18SegAdversarial, SoulyICCV17SegAdversarial} have been also proposed for this problem. Our framework is complimentary to other forms of supervision or adversarial training and can be used alongside them.

    \begin{figure*}[t]
    \begin{center}
    \includegraphics[trim=0 0 0 0,clip, width=0.87\linewidth]{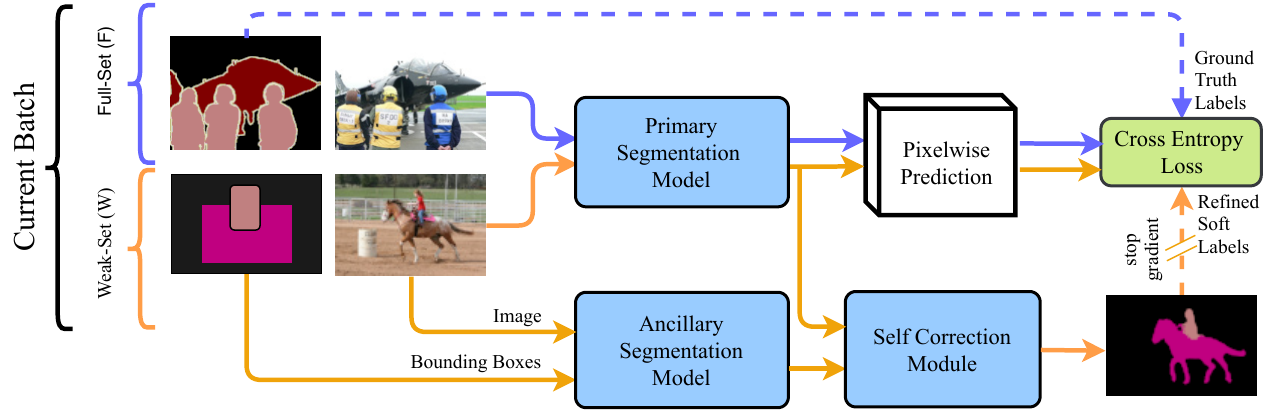}
    \end{center}
    \caption{An overview of our segmentation framework consisting of three models: i) Primary segmentation model generates a semantic segmentation of objects given an image. This is the main model that is subject to the training and is used at test time.
    ii) Ancillary segmentation model outputs a segmentation given an image and bounding box. This model generates an initial segmentation for the weak set, which will aid training the primary model. iii) Self-correction module refines segmentations generated by the ancillary model and the current primary model for the weak set. The primary model is trained using the cross-entropy loss that matches its output to either ground-truth segmentation labels for the fully supervised examples or soft refined labels generated
    by the self-correction module for the weak set.}
    \label{fig_overview}
    \end{figure*}


\vspace{-0.1cm}
\section{Methods}
Our goal is to train a semantic segmentation network in a semi-supervised setting using two training sets: i) a small fully supervised set (containing images, segmentation ground-truth and object bounding boxes) and ii) a weak set (containing images and object bounding boxes only).
An overview of our framework is shown in Fig.~\ref{fig_overview}. There are three models:
i) The \textbf{Primary segmentation model} generates a semantic segmentation of objects given an image. 
ii) The \textbf{Ancillary segmentation model} outputs a segmentation given an image
and bounding box. The model generates an initial segmentation for the weak set,
which aids training of the primary model.
iii) The \textbf{Self-correction module} refines the segmentations generated by the ancillary and current primary model for the weak set.
Both the ancillary and the primary models are based on DeepLabv3+~\cite{ChenECCV18Deeplabv3plus}. 
However, our framework is general and can use any existing segmentation model.

In Sec.~\ref{sec_pseudo_annot}, we present the ancillary model, and in Sec.~\ref{sec:no_self}, we show a simple way to use this model to train the primary model.
In Sec.~\ref{sec:linear_self} and Sec.~\ref{sec:conv_self}, we present two variants of self-correcting model.

\textbf{Notation:} $\x$ represents an image, $\bb$ represents object bounding boxes in an image, and $\y = [\y_1, \y_2, \dots, \y_M]$ represents a segmentation label where $\y_{m} \in [0, 1]^{C+1}$ for $m \in \{1, 2, \dots, M\}$ is a one-hot label for the $m^{th}$ pixel, $C$ is the number of foreground labels augmented with the background class, and
$M$ is the total number of pixels. Each bounding box is associated with an object and has one of the foreground labels. The fully supervised dataset is indicated as 
$\F = \{(\xf, \yf, \bbf)\}_{f=1}^{F}$ where $F$ is the total number of instances
in $\F$. Similarly, the weak set is noted by 
$\W = \{(\xw, \bbw)\}_{w=1}^{W}$. We use $p(\y|\x; \bphi)$ to represent the primary segmentation model
and $p_{anc}(\y|\x, \bb; \btheta)$ to represent the ancillary model. $\bphi$ and $\btheta$
are the respective parameters of each model. We occasionally drop the denotation of parameters for readability. We assume
that both ancillary and primary models define a distribution of
segmentation labels using a factorial distribution, i.e.,
$p(\y|\x; \bphi)=\prod_{m=1}^M p_m(\y_m|\x; \bphi)$ and
$p_{anc}(\y|\x, \bb; \btheta) = \prod_{m=1}^M p_{anc,m}(\y_m|\x, \bb; \btheta)$ where each factor ($p_m(\y_m|\x; \bphi)$ or $p_{anc,m}(\y_m|\x, \bb; \btheta)$) is a categorical distribution (over $C+1$ categories).

\vspace{-0.1cm}
\subsection{Ancillary Segmentation Model} \label{sec_pseudo_annot}
The key challenge in semi-supervised training of
segmentation models with bounding box annotations is to
infer the segmentation of the object inside a box.
Existing approaches to this problem mainly rely on hand-crafted rule-based procedures such as GrabCut~\cite{RotherACM04grabcut} or an iterative label refinement~\cite{PapandreouICCV15ExpMax, DaiICCV15BoxSup, KhorevaCVPR17SimpleDoes} mechanism.
This latter procedure 
typically iterates between segmentation extraction from the image and 
label refinement using the bounding box information 
(for example, by zeroing-out the mask outside of boxes).
The main issues with such procedures are i) bounding box information is not directly used
to extract the segmentation mask, ii) the procedure may be suboptimal as it is hand-designed, 
and iii) the segmentation becomes ambiguous when multiple boxes overlap.

In this paper, we take a different approach by designing an ancillary segmentation model
that forms a per-pixel label distribution given an image and bounding box annotation.
This model is easily trained using the fully supervised set ($\F$)
and can be used as a training signal for images in $\W$.
At inference time, both the image and its bounding box are fed to the network to 
obtain $p_{anc}(\y|\xw, \bbw)$, the segmentation labels distribution. 

\begin{figure}[t]
    \begin{center}
    \includegraphics[trim=0.6cm 16.2cm 9.1cm 1.5cm,clip, width=0.95\linewidth]{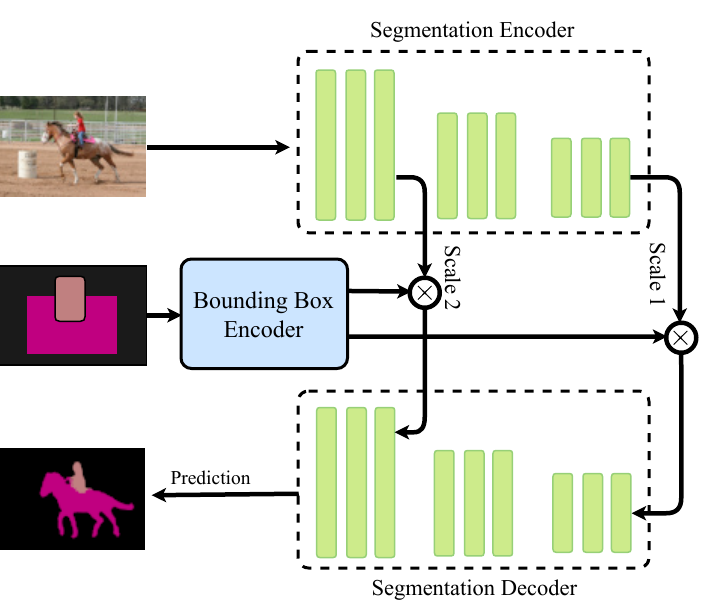}
    \end{center}
    \caption{An overview of the ancillary segmentation model.
    We modify an existing encoder-decoder segmentation model by introducing a bounding box encoder that embeds the 
    box information. The output of the bounding box encoder after passing through a sigmoid activation acts as an attention map.
    Feature maps at different scales from the encoder are fused 
    (using element-wise-multiplication) with attention maps, then passed to the decoder.}
    \label{fig:ancillary}
    \end{figure}

Our key observation in designing the ancillary model 
is that encoder-decoder-based segmentation networks
typically rely on encoders initialized from an image-classification model
(e.g., ImageNet pretrained models). This usually improves the segmentation 
performance by transferring knowledge from
large image-classification datasets. To maintain the same advantage,
we augment an encoder-decoder-based segmentation model with a parallel
\textit{bounding box encoder} network that embeds bounding box information at different scales (See Fig.~\ref{fig:ancillary}). 

The input to the bounding box encoder is a 3D tensor representing a binarized mask of the bounding boxes and a 3D shape representing the target dimensions for the encoder output. The input mask tensor is resized to the target shape then passed through a 3$\times$3 convolution layer with sigmoid activations. The resulting tensor can be interpreted as an attention map which is element-wise multiplied to the feature maps generated by the segmentation encoder.
Fig.~\ref{fig:ancillary} shows two paths of such feature maps at two different scales, as in the DeepLabv3+ architecture. For each scale, an attention map is generated, fused with the corresponding feature map using element-wise multiplication, and fed to the decoder. For an image of size $\mathcal{W}\times \mathcal{H} \times 3$, we represent its object bounding boxes using a binary mask of size $\mathcal{W}\times \mathcal{H} \times(C+1)$ that encodes the $C+1$ binary masks. The $c^{th}$ binary mask at a pixel has the value 1 if it is inside one of the bounding boxes of the $c^{th}$ class. A pixel in the background mask has value 1 if it is not covered by any bounding box. 

The ancillary model is trained using the cross-entropy loss on the full dataset $\F$:
\begin{equation}\label{eq:xent_anc}
\max_{\btheta} \sum_{f\in \F} \log p_{anc}(\yf|\xf, \bbf; \btheta),
\end{equation}
which can be expressed analytically under the factorial distribution assumption. This model is held \textit{fixed} for the subsequent experiments.


\subsection{No Self-Correction} \label{sec:no_self}
We empirically observe that the performance of our ancillary model is 
superior to segmentation models that do not have box information. This is mainly because the bounding box information guides the ancillary model to look for the object inside the box at inference time. 

The simplest approach to training the primary model is to train it to predict using
ground-truth labels on the fully supervised set $\F$ and the labels generated
by the ancillary model on the weak set $\W$. 
For this ``no-self-correction'' model the \textbf{Self-correction module} in Fig.~\ref{fig_overview} merely copies the predictions made by the ancillary segmentation model. Training is guided by optimizing:
\begin{eqnarray}\label{eq:xent_no_corr}
\max_{\bphi} && \sum_{f\in \F} \log p(\yf|\xf; \bphi) + \\
&& \sum_{w \in \W} \sum_{\y} p_{anc}(\y|\xw, \bbw; \btheta) \log p(\y|\xw; \bphi), \nonumber
\end{eqnarray}
where the first term is the cross-entropy loss with one-hot ground-truth labels as target
and the second term is the cross-entropy with soft probabilistic labels generated by
$p_{anc}$ as target. Note that the ancillary model parameterized by $\btheta$ is fixed. We call this approach the
\textit{no self-correction model} as it relies directly on the ancillary model
for training the primary model for examples in $\W$.


\subsection{Linear Self-Correction} \label{sec:linear_self}
Eq.~\ref{eq:xent_no_corr} relies on the ancillary model to predict label distribution
on the weak set. However, this model is trained using only instances of $\F$ without benefit of the data in $\W$. 
Several recent works~\cite{PapandreouICCV15ExpMax, DaiICCV15BoxSup, KhorevaCVPR17SimpleDoes, TanakaCVPR18Noisy, VahdatNIPS17Robust} have incorporated the information in $\W$ by using the primary model itself (as it is being trained on both $\F$ and $\W$) 
to extract more accurate label distributions on $\W$.

Vahdat~\cite{VahdatNIPS17Robust} introduced a regularized Expectation-Maximization algorithm
that uses a linear combination of KL divergences to infer
a distribution over missing labels for general classification problems. 
The main insight is that the inferred distribution $q(\y|\x, \bb)$ over labels
should be close to both the distributions generated by the ancillary model $p_{anc}(\y|\x, \bb)$
\emph{and} the primary model $p(\y|\x)$. However, since the primary model
is not capable of predicting the segmentation mask accurately early in training, 
these two terms are reweighted using a positive scaling factor $\alpha$:
\begin{equation}\label{eq:kl_linear} \small
\min_{q} \KL(q(\y|\x, \bb) || p(\y|\x)) + \alpha \KL(q(\y|\x,\bb) || p_{anc}(\y|\x, \bb)).
\end{equation}
The global minimizer of Eq.~\ref{eq:kl_linear} is obtained as
the weighted geometric mean of the two distributions:
\begin{equation}\label{eq:kl_geo}
q(\y|\x, \bb) \propto \big(p(\y|\x) p_{anc}^{\alpha}(\y|\x, \bb) \big)^{\frac{1}{\alpha+1}}.
\end{equation}
Since both $p_{anc}(\y|\x, \bb)$ and $p(\y|\x)$
decompose into a product of probabilities over the components of $\y$, 
and since the distribution over each component is categorical, then
$q(\y|\x, \bb)=\prod_{m=1}^{M} q_m(\y_m|\x, \bb)$ is also factorial 
where the parameters of the categorical distribution over
each component are computed
by applying softmax activation to the linear combination of
logits coming from primary and ancillary models using
$\sigma\Bigl(\big(\ll_m + \alpha \ \ll_m^{anc}\big)/\big(\alpha + 1\big) \Bigr)$.
Here, $\sigma(.)$ is the softmax function and, 
$\ll_m$ and $\ll_m^{anc}$
are logits 
generated by primary and ancillary models for the $m^{th}$ pixel.

Having fixed $q(\y|\xw, \bbw)$ on the weak set in each iteration of training the primary model,
we can train the primary model using:
\begin{eqnarray}\label{eq:xent_lin_corr}
\max_{\bphi} && \sum_{\F} \log p(\yf|\xf; \bphi) + \\
&& \sum_{\W} \sum_{\y} q(\y|\xw, \bbw) \log p(\y|\xw; \bphi). \nonumber
\end{eqnarray}

Note that $\alpha$ in Eq.~\ref{eq:kl_linear}
controls the closeness of
$q$ to $p(\y|\x)$ and $p_{anc}(\y|\x, \bb)$.
With $\alpha = \infty$, we have
$q=p_{anc}(\y|\x, \bb)$ and the linear self-correction 
in Eq.~\ref{eq:xent_lin_corr} collapses to 
Eq.~\ref{eq:xent_no_corr}, whereas $\alpha = 0$
recovers $q=p(\y|\x)$. A finite $\alpha$
maintains $q$ close to both $p(\y|\x)$ and $p_{anc}(\y|\x, \bb)$. At the beginning of training, $p_{anc}(\y|\x, \bb)$
cannot predict the segmentation label distribution accurately. Therefore, we define a schedule
for $\alpha$ where $\alpha$ is decreased from a large value to a small value 
during training of the primary model.

This corrective model is called the \textit{linear self-correction
model} as it uses the solution to a linear combination of KL divergences (Eq.~\ref{eq:kl_linear}) to infer a distribution over latent segmentation labels.\footnote{In principal, logits of $q_m(\y_m|\x, \bb)$ can be obtained by a \textit{1$\times$1} convolutional layer applied to the depth-wise concatenation of $\ll$ and $\ll^{anc}$  with a fixed averaging kernel. 
This originally motivated us to develop the convolutional self-correction model 
in Sec.~\ref{sec:conv_self} using trainable kernels.}
As the primary model's parameters are optimized during
training, $\alpha$ biases the self-correction mechanism towards the primary model.


\subsection{Convolutional Self-Correction} \label{sec:conv_self}
One disadvantage of linear self-correction is the hyperparameter search
required for tuning the $\alpha$ schedule during training. In this section, we present an approach that overcomes this difficulty by replacing the linear function
with a convolutional network that learns the self-correction mechanism. As a result, the network automatically tunes the mechanism dynamically as the primary model is trained. If the primary model predicts labels accurately, this network can shift its predictions towards the primary model.

\begin{figure}[t]
    \begin{center}
    \includegraphics[trim=0 0 0 0,clip, width=0.99\linewidth]{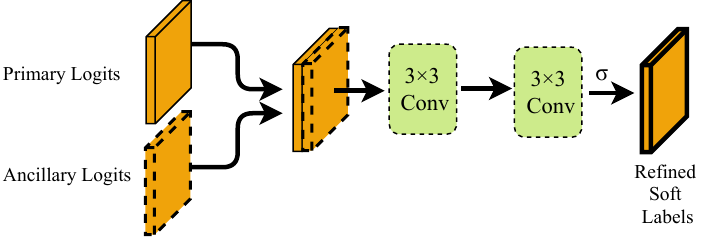}
    \end{center}
    \caption{Convolutional self-correction model learns refining the input label distributions. The subnetwork receives logits from the primary and ancillary models, then concatenates and feeds the output to a two-layer CNN. }
    \label{fig_combine_network}
    \end{figure}

Fig.~\ref{fig_combine_network} shows the architecture of the convolutional self-correcting
model. This small network accepts the logits generated by $p_{anc}(\y|\x, \bb)$
and $p(\y|\x)$ models and generates the factorial distribution $q_{conv}(\y|\x, \bb; \blambda)$ 
over segmentation labels where $\blambda$ represents the parameters of
the subnetwork. The convolutional self-correction subnetwork consists of two convolution layers.  Both layers use a 3$\times$3 kernel and ReLU activations. The first layer has 128 output feature maps and the second has feature maps based on the number of classes in the dataset.

The challenge here is to train this subnetwork such that
it predicts the segmentation labels more accurately than either \pest
or $p(\y|\x)$. To this end, we introduce an additional term in the objective function, 
which trains the subnetwork using training examples in $\F$ while the primary model is being trained on the whole dataset:
\begin{eqnarray}\label{eq:xent_conv_corr}
\max_{\bphi, \blambda} && \sum_{\F} \log p(\yf|\xf; \bphi) + \\
&& \sum_{\W} \sum_{\y} q_{conv}(\y|\xw, \bbw; \blambda) \log p(\y|\xw; \bphi) + \nonumber \\
&& \sum_{\F} \log q_{conv}(\yf|\xf, \bbf; \blambda), \nonumber
\end{eqnarray}
where the first and second terms train the primary model on $\F$ and $\W$
(we do not backpropagate through $q$ in the second term) and the last term
trains the convolutional self-correcting network.

Because the $q_{conv}$ subnetwork is initialized randomly, it is not
able to accurately predict segmentation labels on $\W$ early on during training.
To overcome this issue, we propose the following pretraining procedure:

\begin{enumerate}
    \item Initial training of ancillary model: As with the previous self-correction models, 
    we need to train the ancillary model. Here, half of the fully supervised set ($\F$) is used for this purpose. 
    
    \item Initial training of conv.~self-correction network: The fully supervised data ($\F$) is used to train the primary model and the convolutional self-correcting network. This is done using the first and last terms in Eq.~\ref{eq:xent_conv_corr}.
    
    \item The main training: The whole data ($\F$ and $\W$) are used to \textit{fine-tune} the previous model using the objective function in Eq.~\ref{eq:xent_conv_corr}.
\end{enumerate}

The rationale behind using half of $\F$ in stage 1 is that if we used all $\F$ for training the \pest model, it would train to predict the segmentation mask almost perfectly on this set, therefore, the subsequent training of the convolutional self-correcting network would just learn to rely on \pest. To overcome this training issue, the second half of $\F$ is held out to help the self-correcting network to learn how to combine \pest and $p(\y|\x)$.

\vspace{-0.2cm}
\section{Experiments} \label{sec:exp}
In this section, we evaluate our models on the PASCAL VOC 2012 and Cityscapes datasets. Both datasets contain object segmentation and bounding box annotations. We split the full dataset annotations into two parts to simulate a fully and semi-supervised setting. Similar to ~\cite{ChenECCV18Deeplabv3plus, PapandreouICCV15ExpMax}, performance is measured using the mean intersection-over-union (mIOU) across the available classes. 

{\bf Training: } We use the public Tensorflow~\cite{TensorFlow} implementation of DeepLabv3+~\cite{ChenECCV18Deeplabv3plus} as the primary model. We use an initial learning rate of 0.007 and train the models for 30,000 steps from the ImageNet-pretrained \textit{Xception-65} model~\cite{ChenECCV18Deeplabv3plus}.\footnote{Note that, we do not initialize the parameters from a MS-COCO pretrained model.} For all other parameters we use standard settings suggested by other authors.
At evaluation time, we apply flipping and multi-scale processing for images as in ~\cite{ChenECCV18Deeplabv3plus}. We use 4 GPUs, each with a batch of 4 images. 

We define the following baselines in all our experiments:
\begin{enumerate}
    \item \textbf{Ancillary Model:}  This is the ancillary model, introduced in Sec.~\ref{sec_pseudo_annot}, predicts semantic segmentation labels given an image and its object bounding boxes. This model is expected to perform better than other models as it uses bounding box information. 
    
    \item \textbf{No Self-correction:} This is the primary model trained using
    the model introduced in Sec.~\ref{sec:no_self}.
    
    \item \textbf{Lin.~Self-correction:} This is the primary model
    trained with linear self-correction as in Sec.~\ref{sec:linear_self}.
    
    \item \textbf{Conv.~Self-correction:} The primary model trained with the convolutional self-correction as in Sec.~\ref{sec:conv_self}.
    
    \item \textbf{EM-fixed Baseline:} Since our linear self-correction model is derived
    from a regularized EM model~\cite{VahdatNIPS17Robust}, we compare our model
    with Papandreou \etal~\cite{PapandreouICCV15ExpMax} which is also an EM-based model. 
    We implemented their EM-fixed baseline with DeepLabv3+ for fair comparison.
    This baseline achieved the best results in \cite{PapandreouICCV15ExpMax} for semi-supervised learning.

\end{enumerate}

For linear self-correction, $\alpha$ controls the weighting in the KL-divergence bias with large $\alpha$ favoring the ancillary model and small $\alpha$ favoring the primary model. We explored different starting and ending values for $\alpha$ with an exponential decay in-between. We find that a starting value of $\alpha=30$ and the final value of $\alpha=0.5$ performs well for both datasets. This parameter setting is robust as moderate changes of these values have little effect. 

\subsection{PASCAL VOC Dataset}
In this section, we evaluate all models on the PASCAL VOC 2012 segmentation benchmark~\cite{Pascal15}. This dataset consists of 1464 training, 1449 validation, and 1456 test images covering 20 foreground object classes and one background class for segmentation. An auxiliary dataset of 9118 training images is provided by~\cite{BharathICCV11ExtraPascalSeg}. We suspect, however, that the segmentation labels of ~\cite{BharathICCV11ExtraPascalSeg} contain a small amount of noise. In this section, we refer to the union of the original PASCAL VOC training dataset and the auxiliary set as the \textit{training} set.
We evaluate the models mainly on the validation set and the best model is evaluated only once on the test set using the online evaluation server. 

\begin{figure*}[htp]
    \vspace{-0.4cm}
    \begin{tabularx}{\columnwidth}{cccccc}

    	\vspace{-0.45cm}
		\subfloat{\photo{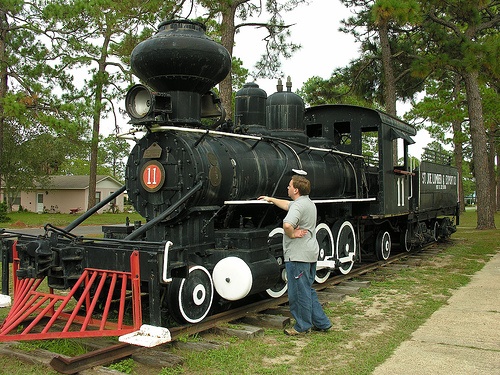}}         & 
		\subfloat{\photo{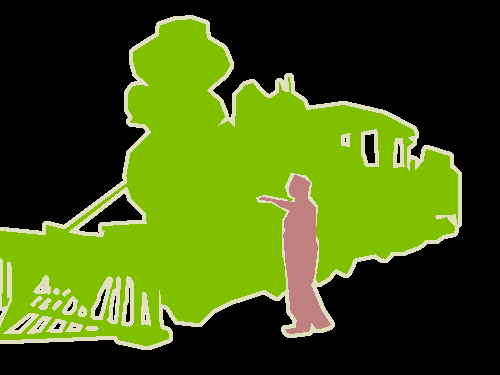}}         & 
		\subfloat{\photo{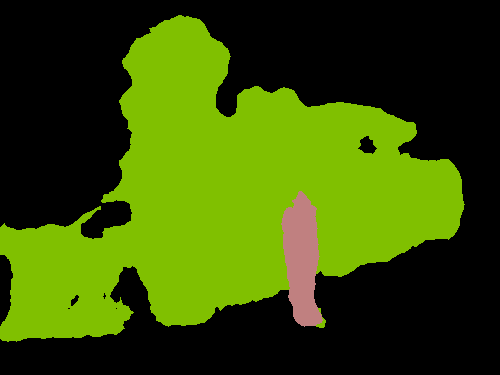}}         & 
		\subfloat{\photo{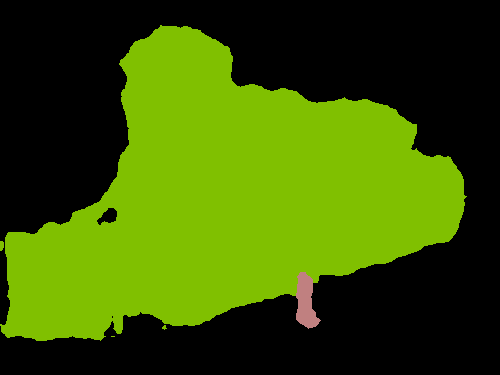}}         & 
		\subfloat{\photo{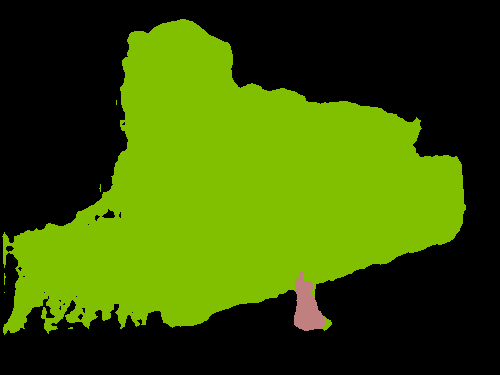}}         & 
		\subfloat{\photo{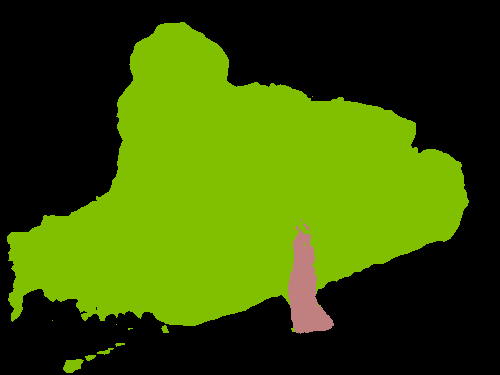}}        \\
	    
	    \vspace{-0.45cm}
		\subfloat{\photo{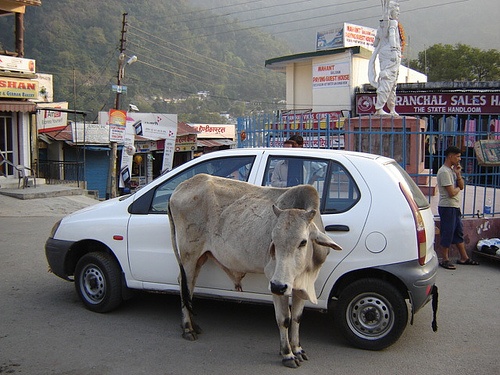}}         & 
		\subfloat{\photo{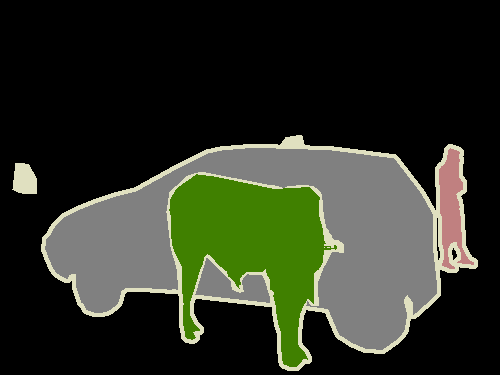}}         & 
		\subfloat{\photo{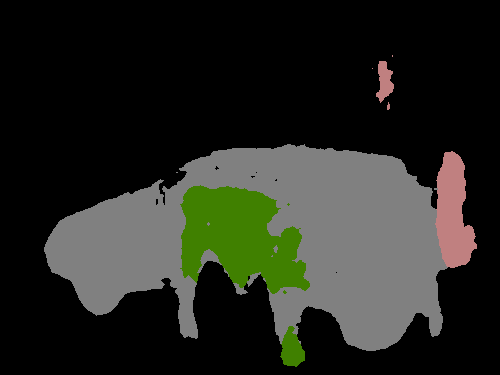}}         & 
		\subfloat{\photo{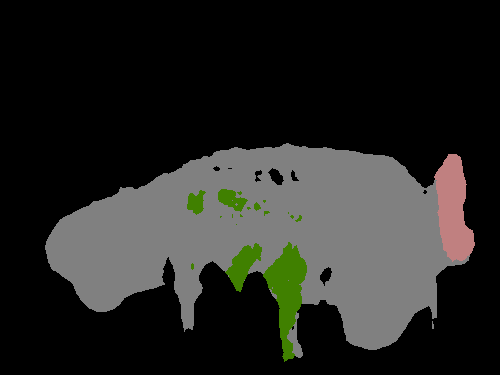}}         & 
		\subfloat{\photo{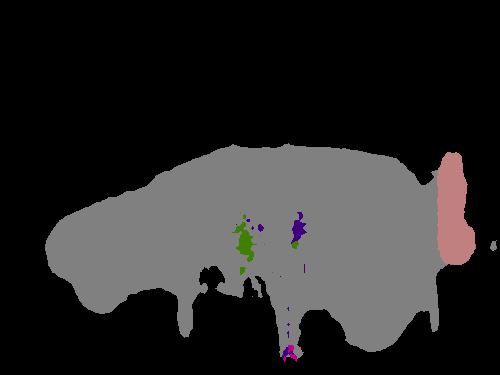}}         & 
		\subfloat{\photo{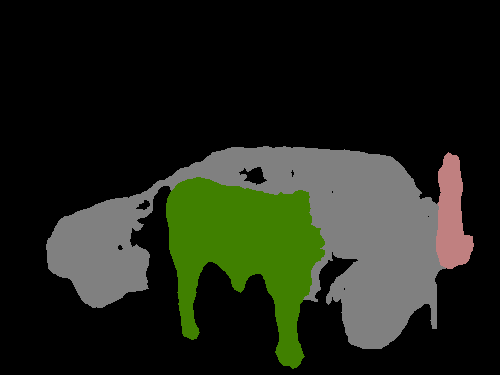}}        \\	
	
	    \vspace{-0.45cm}
		\subfloat{\photo{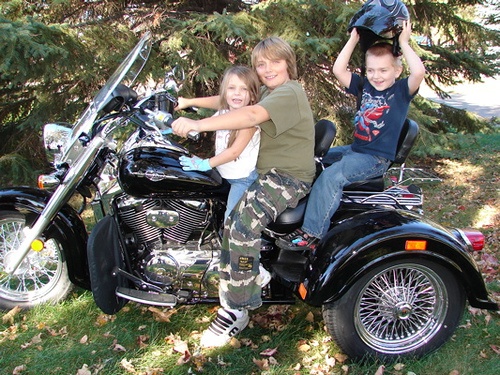}}         & 
		\subfloat{\photo{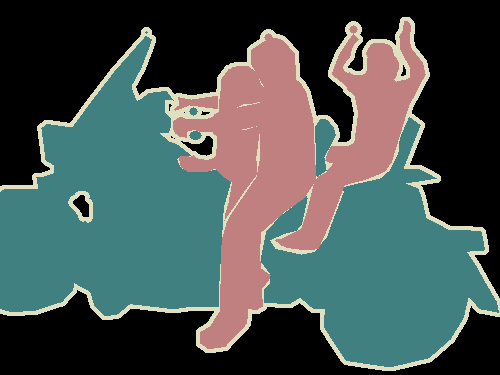}}         & 
		\subfloat{\photo{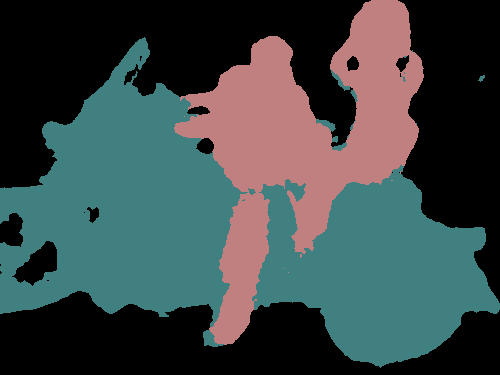}}         & 
		\subfloat{\photo{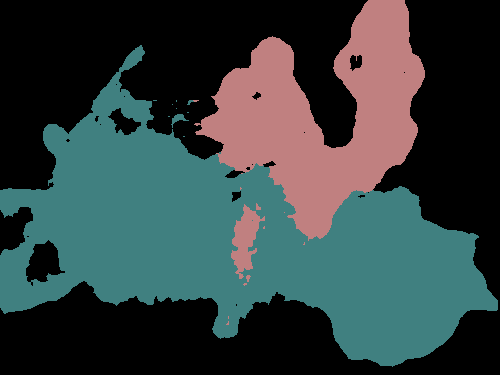}}         & 
		\subfloat{\photo{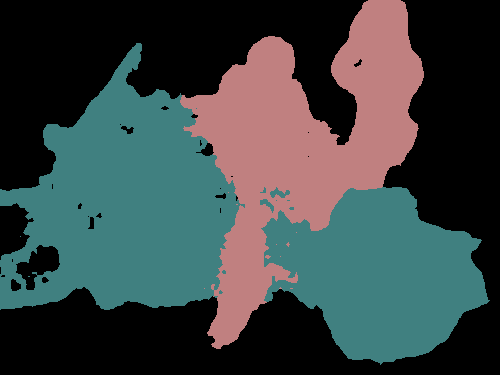}}         & 
		\subfloat{\photo{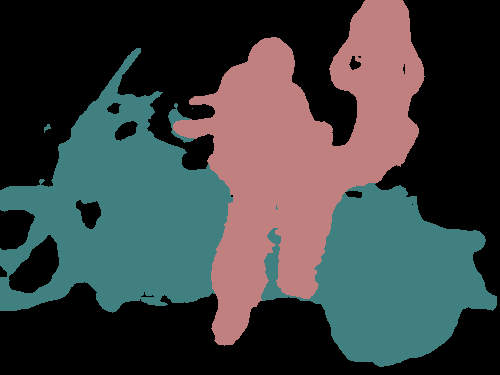}}        \\	
	
		
	    \vspace{-0.01cm}
		\subfloat{\photo{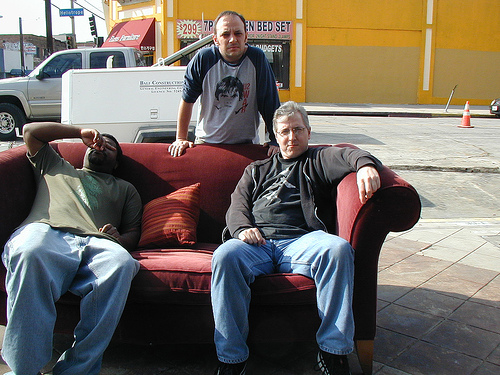}}         & 
		\subfloat{\photo{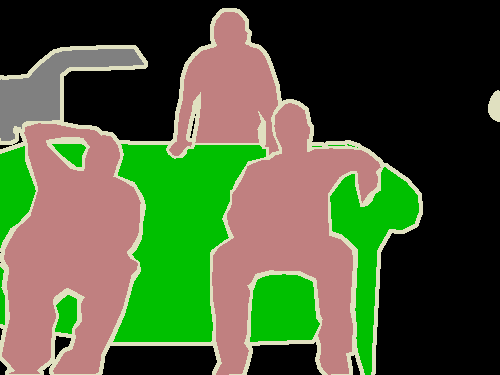}}         & 
		\subfloat{\photo{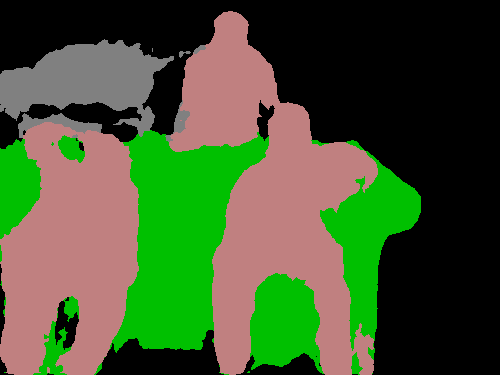}}         & 
		\subfloat{\photo{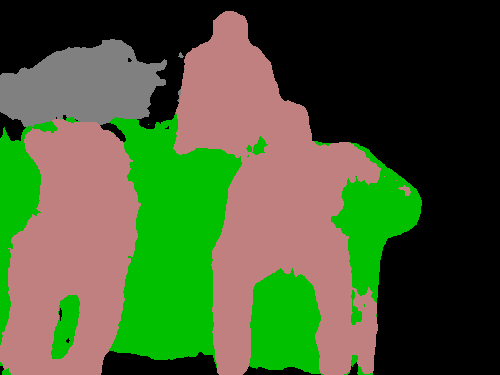}}         & 
		\subfloat{\photo{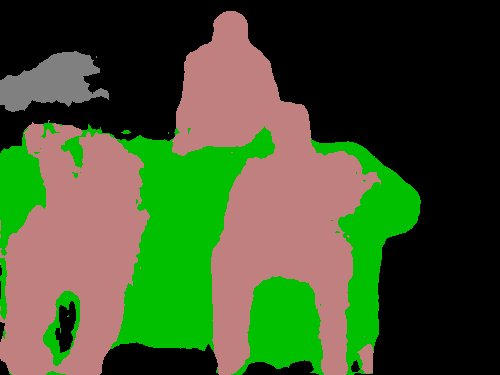}}         & 
		\subfloat{\photo{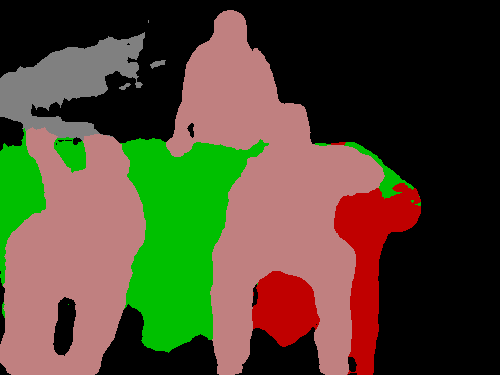}}      \\
		
		Input Image     &   ground-truth    &   Ancillary Model &    No Self-correction  &   Lin.~Self-correction       &   Conv.~Self-correction 
		
    \end{tabularx}
    \caption{Qualitative results on the \textbf{PASCAL VOC 2012 validation} set. The last four columns represent the models in column 1464 of Table~\ref{table_pascal_est_alpha}. The Conv.~Self-correction model typically segments objects better than other models. }
    \label{vis_pascal_main}
    \end{figure*}

In Table~\ref{table_pascal_est_alpha}, we show the performance of different variants of our model for different sizes of the fully supervised set $\F$. The remaining examples in the training set are used as $\W$. We make several observations from  Table~\ref{table_pascal_est_alpha}: 
i) The ancillary model that predicts segmentation labels given an image and its object bounding boxes performs well even when it is trained with a training set as small as 200 images. This shows that this model can also provide a good training signal for the weak set that lacks segmentation labels.
ii) The linear self-correction model typically performs better than no self-correction model supporting our idea that combining the primary and ancillary model for inferring segmentation labels results in better training of the primary model.
iii) The convolutional self-correction model performs comparably or better than the linear self-correction while eliminating the need for defining an $\alpha$ schedule.  Fig.~\ref{vis_pascal_main} shows the output of these models.   

\setlength{\tabcolsep}{1pt}
\begin{table}[ht]
\begin{center}
\begin{tabular}{|c|>{\centering}m{0.9cm}|>{\centering}m{0.9cm}|>{\centering}m{0.9cm}|c|}
\hline
\# images in $\F$ & 200         &   400         &   800         &   1464        \\
\hline
Ancillary Model       & 81.57       &   83.56       &  85.36        &    86.71      \\
\hline
No Self-correction    & 78.75       &   79.19       &  80.39        &    80.34      \\ 
Lin.~Self-correction  & \bf 79.43   &   79.59       &  \bf 80.69    &    81.35      \\
Conv.~Self-correction & 78.29       &   \bf 79.63   &  80.12        &    \bf 82.33       \\
\hline
\end{tabular}
\end{center}
\caption{Ablation study of models on the \textbf{PASCAL VOC 2012 validation} set using mIOU for different sizes of $\F$. For the last three rows, the remaining images in the training set is used as $\W$, i.e. $W + F = 10582$.}
\label{table_pascal_est_alpha}
\end{table}

Table~\ref{table_pascal_mix} compares the performance of our models against different baselines and published results. In this experiment, we use 1464 images as $\F$
and 9118 images originally from the auxiliary dataset as $\W$. Both self-correction models achieve similar results and outperform other models. 

\begin{figure*}[htp]
    \begin{tabularx}{\columnwidth}{cccccc}
	    \vspace{-0.45cm}
		\subfloat{\photo{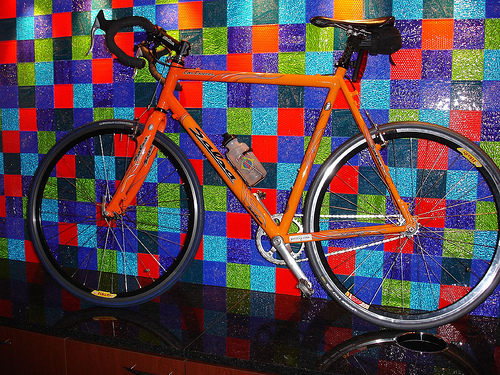}}                & 
		\subfloat{\photo{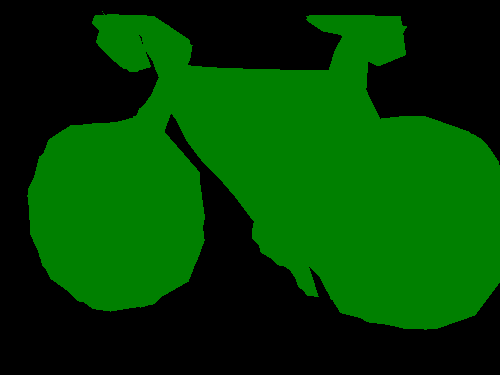}}                & 
		\subfloat{\photo{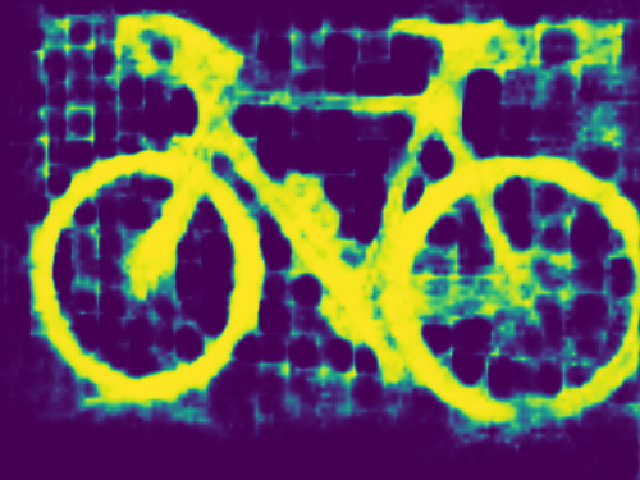}}          & 
		\subfloat{\photo{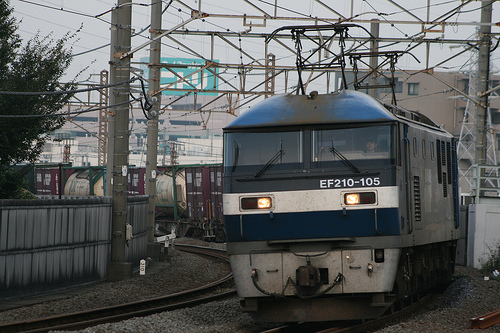}}                & 
		\subfloat{\photo{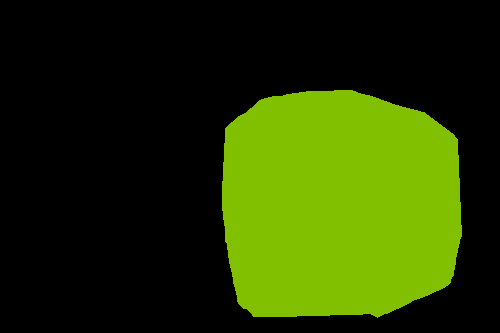}}                & 
		\subfloat{\photo{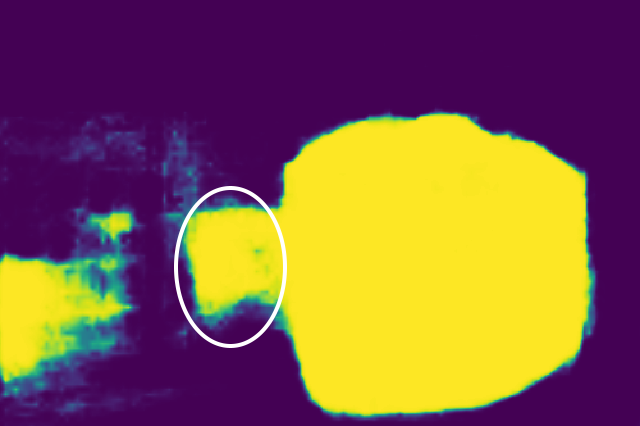}}       \\
		
	    \vspace{-0.45cm}
		\subfloat{\photo{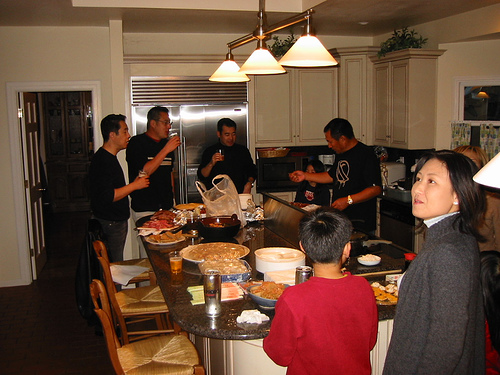}}                & 
		\subfloat{\photo{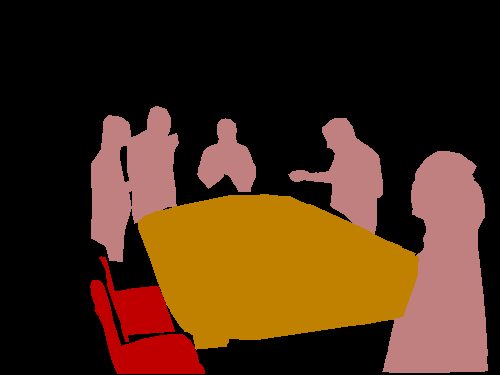}}                & 
		\subfloat{\photo{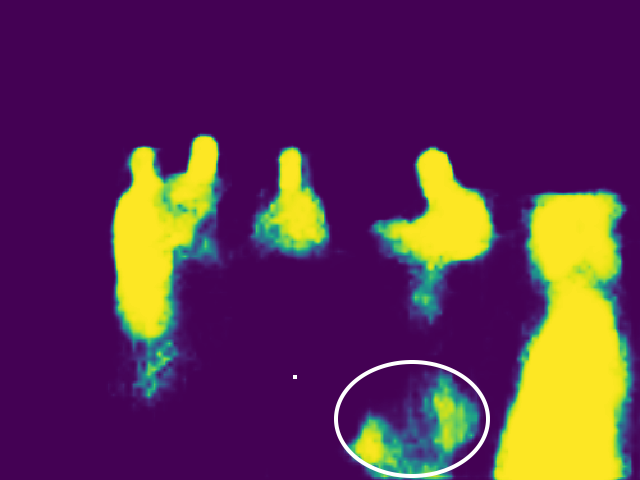}}       & 
		\subfloat{\photo{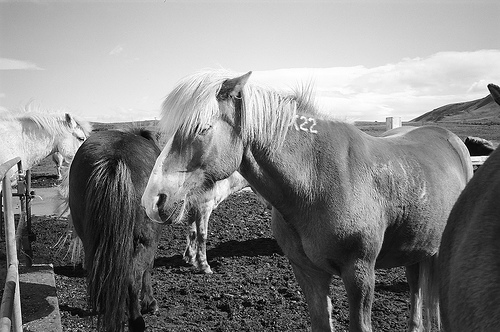}}                & 
		\subfloat{\photo{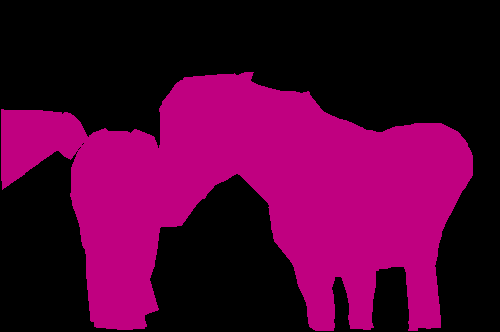}}                & 
		\subfloat{\photo{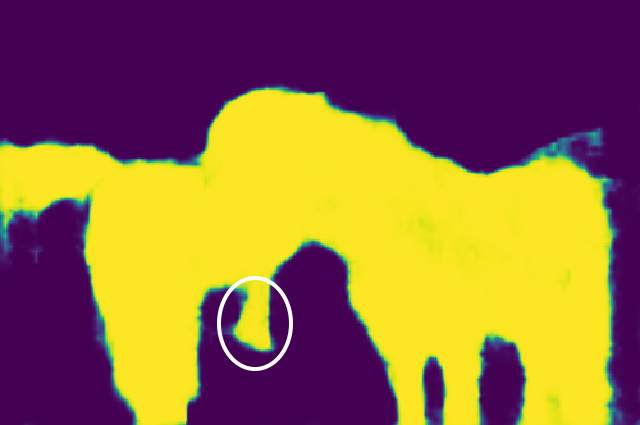}}       \\
		\vspace{-0.01cm}
		\subfloat{\photo{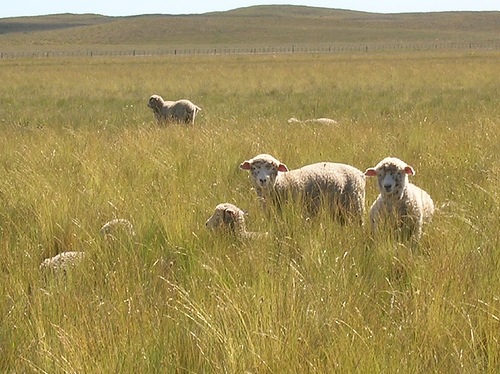}}                & 
		\subfloat{\photo{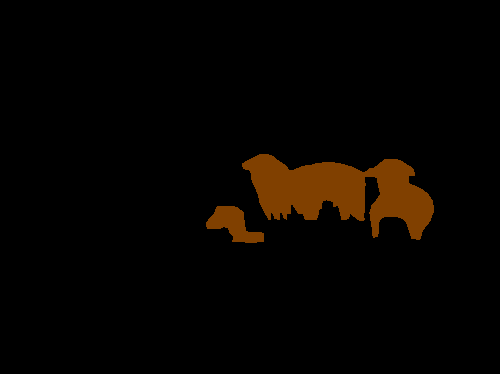}}                & 
		\subfloat{\photo{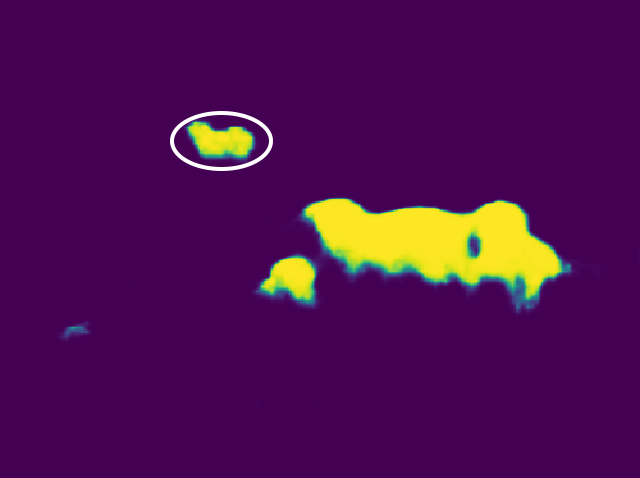}}       & 
		\subfloat{\photo{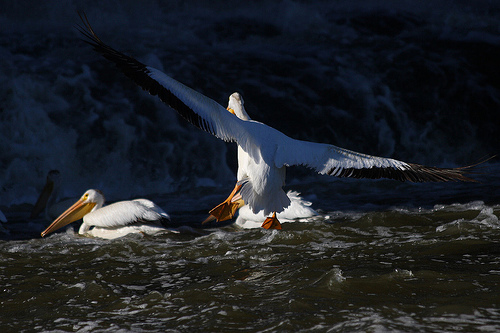}}                & 
		\subfloat{\photo{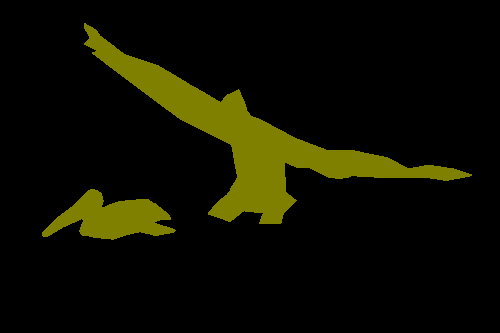}}                & 
		\subfloat{\photo{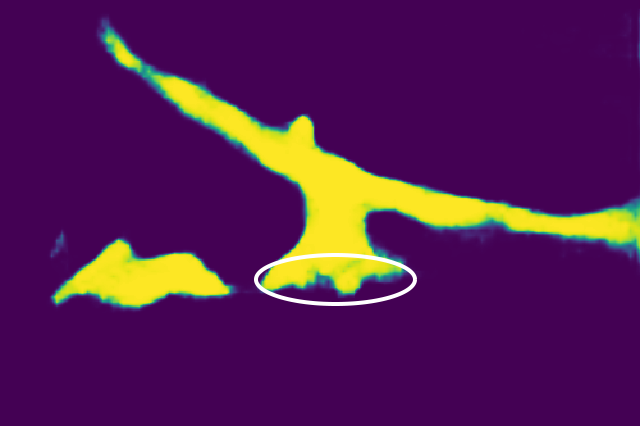}}       \\      
		Input Image     &   Ground-truth    &   Ancillary Heatmap &    Input Image     &   Ground-truth    &   Ancillary Heatmap 
		
    \end{tabularx}
    \caption{Qualitative results on the \textbf{PASCAL VOC 2012 auxiliary} (the weak set). The heatmap of a single class for the ancillary model is shown for several examples.
    The ancillary model can successfully correct the labels for 
    \textit{missing} or \textit{over-segmented} objects in these images (marked by ellipses).}
    \label{vis_pascal_ancilary}
    \end{figure*}

\textbf{Surprisingly}, our semi-supervised models outperform the fully supervised model. We hypothesize two possible explanations for this observation. Firstly, this may be due to label noise in the 9k auxiliary set~\cite{BharathICCV11ExtraPascalSeg} that negatively affects performance of Vanilla DeepLapv3+. As evidence, Fig.~\ref{vis_pascal_ancilary} compares the output of the ancillary model with ground-truth annotations and highlights some of improperly labeled instances. Secondly, the performance gain may also be due to explicit modeling of label uncertainty and self-correction. To test this hypothesis, we train vanilla DeepLabv3+ on only 1.4K instances in the original PASCAL VOC 2012 training set\footnote{The auxiliary set is excluded to avoid potentially noisy labels.} and obtain 68.8\% mAP on the validation set. However, if we train the convolutional self-correction model on the same training set and allow the model to \textit{refine} the ground truth labels using self-correction\footnote{For this experiment 1.1K images are used as $\F$ and 364 images as $\W$. For $\W$, we let self-correction model to refine the original ground-truth labels.}, we get mAP as high as 76.88\% (the convolutional self correction on top of bounding boxes yields 75.97\% mAP). This indicates that modeling noise with robust loss functions and allowing for self-correction may significantly improve the performance of segmentation models. This is consonant with self-correction approaches that have been shown to be effective for edge detection~\cite{yu2018seal, acuna2019devil}, and is in contrast to common segmentation objectives which train models using cross-entropy with one-hot annotation masks. Very similar to our approach and reasoning, ~\cite{Feng2019PoseEstimation} uses logits to train a lightweight pose estimation model using knowledge distillation technique.

Unfortunately, the state-of-the-art models are still using the older versions of DeepLab. It was infeasible for us to either re-implement most of these approaches using DeepLabv3+ or re-implement our work using old versions. The only exception is EM-fixed baseline~\cite{PapandreouICCV15ExpMax}. Our re-implementation using DeepLabv3+ achieves 79.25\% on the validation set while the original paper has reported 64.6\% using DeepLabv1. In the lower half of Table~\ref{table_pascal_mix}, we record previously published results (using older versions of DeepLab). A careful examination of the results show that \textit{our work is superior} to previous work as our semi-supervised models outperform the fully supervised model while previous work normally do not.

Finally, comparing Table~\ref{table_pascal_est_alpha} and \ref{table_pascal_mix}, we see that with $F=200$ and $W=10382$, our linear~self-correction model performs similarly to DeepLabv3+ trained with the whole dataset. Using the labeling cost reported in \cite{BearmanECCV16WeakPoint}, this theoretically translates to a $\sim$7x reduction in annotation cost.

  \setlength{\tabcolsep}{1pt}
  \begin{table}[ht]
  \begin{center}
  \begin{tabular}{|c|c|c|c|c|}
  \hline
  \multicolumn{2}{|c|}{Data Split} & \multirow{2}{*}{Method} & \multirow{2}{*}{Val} & \multirow{2}{*}{Test}  \\
  \cline{1-2}
  $F$ & $W$ & & & \\
  \hline
  1464 & 9118 & No Self-Corr.                             & 80.34 & 81.61  \\
  1464 & 9118 & Lin. Self-Corr.                            & 81.35 & 81.97 \\
  1464 & 9118 & Conv. Self-Corr.                           & \bf 82.33 & \bf 82.72 \\
  1464 & 9118 & EM-fixed Ours~\cite{PapandreouICCV15ExpMax}     & 79.25 & - \\
  10582 & -   & Vanilla DeepLabv3+~\cite{ChenECCV18Deeplabv3plus}                           & 81.21 & - \\
  \hline \hline
  1464 & 9118 & BoxSup-MCG~\cite{DaiICCV15BoxSup}           & 63.5 & -\\
  1464 & 9118 & EM-fixed~\cite{PapandreouICCV15ExpMax}      & 65.1 & -\\
  1464 & 9118 & M $\cap$ G+~\cite{KhorevaCVPR17SimpleDoes}  & 65.8 & -\\
    1464 & 9118 & FickleNet~\cite{lee2019ficklenet}  & 65.8 & -\\
    1464 & 9118 & Song \etal~\cite{Chunfeng2019SemSeg}  & 67.5 & -\\
    10582 & - & Vanilla DeepLabv1~\cite{ChenICLR15DeeplabV1} & 69.8 & - \\
  \hline
  \end{tabular}
  \end{center}
  \caption{Results on \textbf{PASCAL VOC 2012 validation and test} sets. The last three rows report the performance of previous semi-supervised models with the same annotation. 
  }
  \label{table_pascal_mix}
  \end{table}

\setlength{\tabcolsep}{1pt}
\begin{table}[h]
\begin{center}
\begin{tabular}{|c|>{\centering}m{0.9cm}|>{\centering}m{0.9cm}|c|}
\hline
\# images in $\F$     & 200         &   450         &   914       \\
\hline
Ancillary Model       & 79.4        &   81.19       &  81.89     \\
\hline
No Self-correction    & \bf 73.69   &   75.10       &  75.44      \\ 
Lin.~Self-correction  & 73.56       &   75.24       &  76.22      \\
Conv.~Self-correction & 69.38       &   \bf 77.16   &  \bf 79.46       \\
\hline
\end{tabular}
\end{center}
  \caption{Ablation study of our models on \textbf{Cityscapes validation} set using mIOU for different sizes of $\F$. For the last three rows, the remaining images in the training set are used as $\W$, i.e., $W + F = 2975$.}
  \label{table_city_est_alpha}
\end{table}

  \setlength{\tabcolsep}{1pt}
  \begin{table}
  \begin{center}
  \begin{tabular}{|c|c|c|c|}
  \hline
  \multicolumn{2}{|c|}{Data Split} & \multirow{2}{*}{Method} & \multirow{2}{*}{mIOU}   \\
  \cline{1-2}
  $F$ & $W$ & & \\
  \hline
  914 & 2061 & No Self-Corr.                             & 75.44  \\
  914 & 2061 & Lin. Self-Correction                        & 76.22 \\
  914 & 2061 & Conv. Self-Correction                       & \bf 79.46 \\
  914 & 2061 &  EM-fixed~\cite{PapandreouICCV15ExpMax}      & 74.97  \\
  2975 & -  & Vanilla DeepLabv3+$_{ours}$                            & 77.49 \\
  \hline
  \end{tabular}
  \end{center}
  \caption{Results on \textbf{Cityscapes validation} set. 30\% of the training examples is used as $\F$, and the remaining as $\W$.}
  \label{table_city_mix}
  \end{table}

\subsection{Cityscapes Dataset}
In this section we evaluate performance on the Cityscapes dataset~\cite{Cityscapes16} which contains images collected from cars driving in cities during different seasons. This dataset has good quality annotations, however some instances are over/under segmented. It consists of 2975 training, 500 validation, and 1525 test images covering 19 foreground object classes (stuff and object) for the segmentation task. However, 8 of these classes are flat or construction labels (e.g., road, sidewalk, building, and etc.), and a very few bounding boxes of such classes cover the whole scene. To create
an object segmentation task similar to the PASCAL VOC dataset, we use only 11 classes (pole, traffic light, traffic sign, person, rider, car, truck, bus, train, motorcycle, and bicycle) as foreground classes and all other classes are assigned as background. Due to this modification of labels, we report the results only on the validation set, as the test set on server evaluates on all classes. We do not use the coarse annotated training data in the dataset.

Table~\ref{table_city_est_alpha} reports the performance of our model for an increasing
number of images as $\F$, and Table~\ref{table_city_mix}
compares our models with several baselines similar to the previous dataset. The same conclusion and insights observed on the PASCAL dataset hold for the Cityscapes dataset indicating the efficacy of our self-corrective framework.


\section{Conclusion}
In this paper, we have proposed a semi-supervised framework for training deep CNN segmentation models using a small set of fully labeled
and a set of weakly labeled images (boxes annotations only). We introduced two mechanisms that enable the underlying primary model to correct the weak labels
provided by an ancillary model. The proposed self-correction mechanisms combine the predictions made by the primary and ancillary model
either using a linear function or trainable CNN. The experiments show that our proposed framework outperforms previous semi-supervised models
on both the PASCAL VOC 2012 and Cityscapes datasets. Our framework can also be applied to the instance segmentation task~\cite{HuCVPR18SegmentEveryThing, ZhouCVPR18WeakPeakResponse, ZhaoCVPR18WeakPseudoMask}, but we leave further study of this to future work.

{\small
\bibliographystyle{ieee_fullname}
\bibliography{egpaper_for_review}
}

\end{document}